\documentclass{article}
\usepackage[final]{nips_2018}
\usepackage[]{nips_2018}

\usepackage[utf8]{inputenc} 
\usepackage[T1]{fontenc}    
\usepackage{hyperref}       
\usepackage{url}            
\usepackage{booktabs}       
\usepackage{amsfonts}       
\usepackage{nicefrac}       
\usepackage{microtype}      
\usepackage{hyperref}
\usepackage{enumerate}
\pdfoutput=1
\usepackage{multirow}   
\usepackage{bbm}
\usepackage{graphicx}
\usepackage{algorithm}
\usepackage{algorithmic}
\usepackage{natbib}
\usepackage{subcaption}
\usepackage{caption}
\usepackage{placeins}
\usepackage{amsmath}
\usepackage{mathtools}
\usepackage{xcolor}
\usepackage{braket}

\newcommand{\code}[1]{\texttt{#1}}

\newcommand*\samethanks[1][\value{footnote}]{\footnotemark[#1]}
\newcommand{\mX}{\mathcal{X}}
\newcommand{\mY}{\mathcal{Y}}

\title{Accurate, Data-Efficient Learning from Noisy, Choice-Based Labels for Inherent Risk Scoring}
\author{
  W. Ronny Huang\thanks{Equal contribution.} \\
  Ernst \& Young LLP\\
  New York, NY \\
  \texttt{ronny.huang@ey.com} \\
  \And
  Miguel A. Perez\samethanks \\
  Ernst \& Young LLP\\
  New York, NY \\
  \texttt{miguel.a.perez@ey.com} \\
}

\begin{document}
\maketitle
\begin{abstract}
  Inherent risk scoring is an important function in anti-money laundering, used for determining the riskiness of an individual during onboarding before fraudulent transactions occur. It is, however, often fraught with two challenges: (1) inconsistent notions of what constitutes as high or low risk by experts and (2) the lack of labeled data. This paper explores a new paradigm of data labeling and data collection to tackle these issues. The data labeling is choice-based; the expert does not provide an absolute risk score but merely chooses the most/least risky example out of a small choice set, which reduces inconsistency because experts make only relative judgments of risk. The data collection is synthetic; examples are crafted using optimal experimental design methods, obviating the need for real data which is often difficult to obtain due to regulatory concerns. We present the methodology of an end-to-end inherent risk scoring algorithm that we built for a large financial institution. The system was trained on a small set of synthetic data (188 examples, 24 features) whose labels are obtained via the choice-based paradigm using an efficient number of expert labelers. The system achieves 89\% accuracy on a test set of 52 examples, with an area under the ROC curve of 93\%.
\end{abstract}

\section{Introduction}

Financial institutions today are tasked with Know Your Customer obligations in order to mitigate money laundering activity in their systems that often serves to enable terrorism or drug trafficking. Banks as a result have collected large amounts of background information about their customers, such as their source of funds, business operations, and financial statements. Accurately predicting customer {\em inherent risk} from this information is a critical function in preventing money laundering and other fraudulent behaviors {\em before} they happen.

Despite the abundance of customer information (features), there is a paucity of data on actual instances of financial crime (labels). Traditionally, expert money laundering investigators are employed to sift through large amounts of customer records and provide their judgments on (i.e., act as a prediction module for) the risk level of each individual. Because such a process is costly, manual, and time-consuming, much work is focused on building accurate machine learning models for predicting inherent risk.

Inherent risk is a concept that most find easy to evaluate in {\em comparisons}. For example, one might easily judge that a politically exposed person with frequent transactions of more than \$10k to foreign parties is inherently riskier {\em compared to} a customer with smaller, more stable transactions, who in turn is inherently riskier compared to a long-established domestic customer on a fixed-term deposit. However, due to the subjective and ever-changing nature of the notion of risk, it is far more difficult to judge the inherent risk of an individual on a standard scale, e.g., between -1 and 1, than it is to make comparisons between different examples. In practice, investigators are often asked only to provide labels on a binary scale of either ``high'' or ``low'' risk. While such binning strategies may make it easier for investigators, they also allow for highly inconsistent and noisy labels, because different investigators have different ideas of what constitutes as, e.g., ``high'' risk.

\subsection{Contributions}

To address the noisy and subjective nature of inherent risk labeling, we present a novel method for obtaining standard scale continuous valued labels of inherent risk from purely choice-based queries from a crowd of expert labelers. Contributions include the following: (1) leveraging a Monte Carlo D-optimal design-of-experiments approach for generating a set of synthetic customer examples which covers the input space without bias; (2) an optimal algorithm for generating choice sets which minimizes redundant pairings; and (3) a formulation which aggregates choice-based queries into continuous target values with maximal, unbiased use of the information provided. Finally, we show that our end-to-end algorithm learned from such labels achieves a classification accuracy of 89\% on a test set of customers. This is, to the authors' knowledge, the first time choice-based conjoint analysis techniques have been successfully applied to financial crimes risk prediction.

\begin{figure}[!t]
\vskip -.4cm
\centering
\begin{subfigure}[b]{.35\textwidth}
    \centering
    \includegraphics[width=\textwidth]{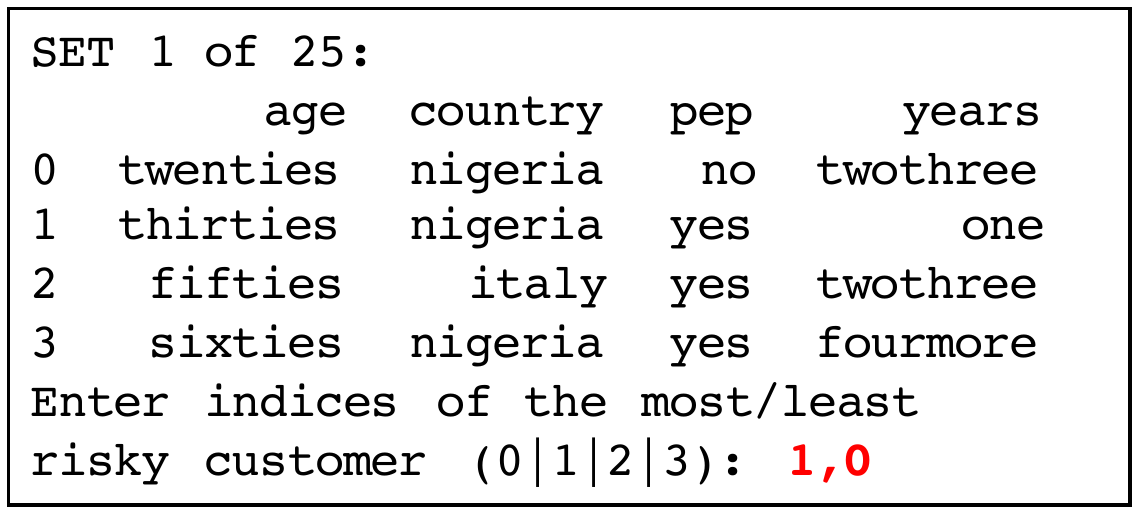}
    \caption{Example of a choice set.}
    \label{fig:question}
\end{subfigure}
\qquad
\begin{subfigure}[b]{.4\textwidth}
    \centering
    \includegraphics[width=\textwidth]{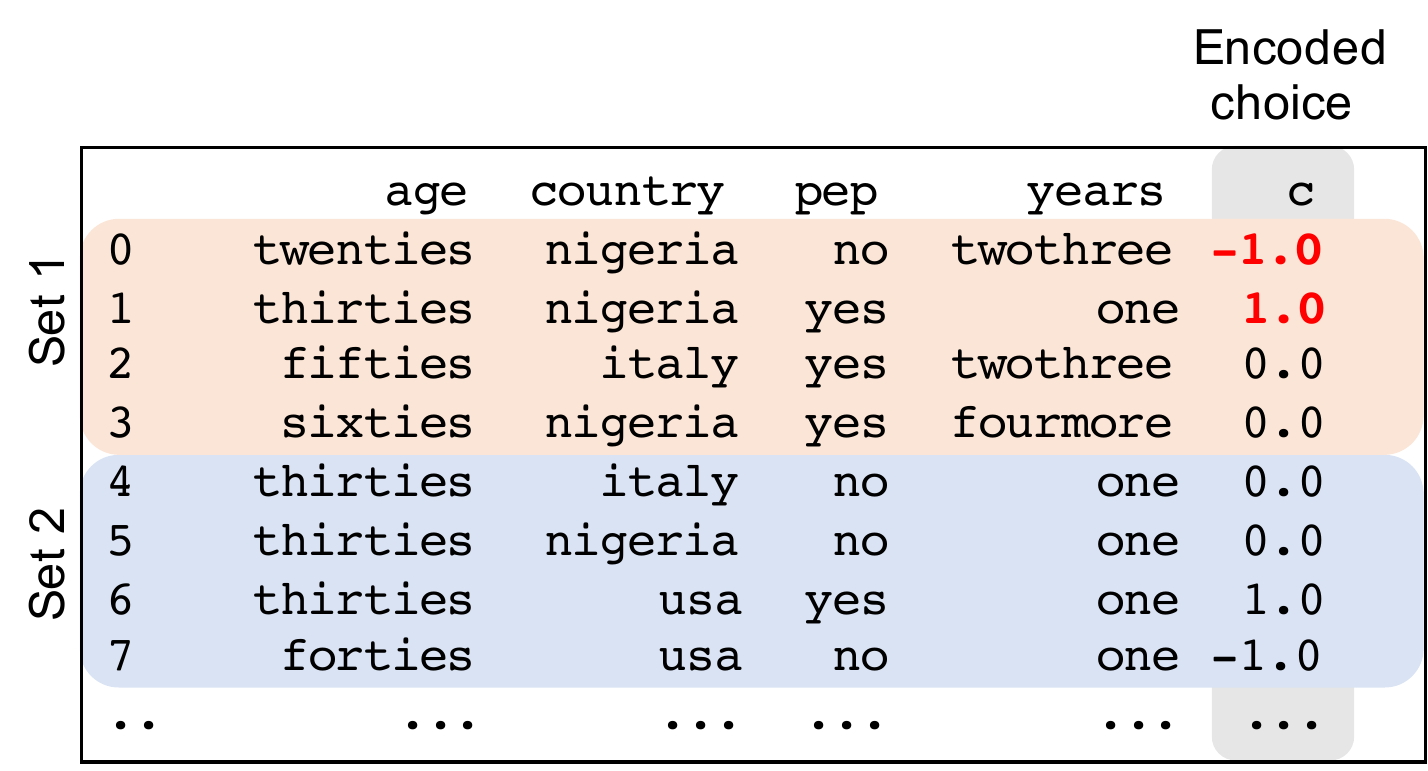}
    \caption{Example data layout.}
    \label{fig:datalayout}
\end{subfigure}
\caption{\small Illustration of the choice-based format for querying the experts. (a) Experts are presented with a small minibatch (choice set size of 4 here) of examples, and they are asked to select only the highest and lowest risk example from within that set. (b) Layout of the data obtained after labeling of a choice-based questionnaire. Most/least risky examples from each choice set are encoded as 1/-1 while unselected examples are encoded as 0. These encodings are mean-aggregated over multiple questionnaires (with different minibatching) to obtain the final label. See Section \ref{sec:choice2score} for more discussion.}
\label{fig:choiceexample}

\end{figure}

\section{The Choice-Based Labeling Paradigm}

\subsection{Conjoint Analysis: A questionnaire of Label Querying Formats}
We first motivate the need for choice-based label querying. Conjoint analysis is a method wherein the responses from one or more human labelers (who have access to the oracle labeling function with some error, i.e., $y=f(x)+\epsilon$) are used to label a set of examples for supervised training. The labels can be queried in several formats, including direct, ranking, or choice formats. Table \ref{tab:conjoint} summarizes their properties.

The direct label format is typically assumed in most machine learning formulations (i.e., $y_i$ is directly queried for every example $x_i$). However, in risk scoring, the expert often has a relative notion of the true label and is unable to provide $y_i$ directly. Binning the label value into discrete ordinal values (e.g., high/medium/low) is one solution but still suffers from subjectivity.

In contrast, the ranking and choice formats demand less of the expert---they require only pairwise comparisons (e.g., Is example A riskier than example B?) which alleviate the need for labeling on a standard, absolute scale. Such labeling formats are often employed in marketing to learn customer preferences on variations of potential products, where again the expert, i.e., customer, has only a relative notion of the true label, i.e., utility of one product vs. another (\cite{Asioli2016}). Table \ref{tab:conjoint} shows a comparison of the three formats.

The ranking format asks the expert to sort the entire set of examples according to their target value (risk score in our case), a task that can be broken down into a series of pairwise comparisons. However, it is time-consuming for a human expert to sort hundreds or thousands of examples---using MergeSort, one could achieve $O(n\log n)$ time complexity at best. In the choice format, the expert is presented with small choice sets, or minibatches, of examples and asked only to select the most and least risky examples from within that set. In addition to having a time complexity of $O(n)$, choice-based formats tend to be most natural for human experts (\cite{sethuraman2005}) because human experts can ``eye-ball'' the choice set and quickly determine the examples at the extremes of the risky/non-risky spectrum, without needing to deliberate between the relative risk levels of those examples in between. For these reasons, the choice-based format is most popular in conjoint analysis and is the format that we will henceforth consider. Figure \ref{fig:question} shows an example of a choice set, and Figure \ref{fig:datalayout} shows an example of how choice-based information queried from the human experts is stored.

\begin{table}[b]
\caption{Comparison and descriptions of three formats of label querying.}
\label{tab:conjoint}
\begin{center}
\begin{small}
\begin{tabular}{llc}
\hline
Format & Information obtained & Mathematical expression \\
\hline
Direct & Continuous valued labels & $y_i, \forall i$ \\
Ranking & Ranking of all labels & $\{r(i), \forall i \mid y_i\! >\! y_j, \forall\: r(i)\! >\! r(j) \}$ \\
Choice & Max/min label w/in choice set & $\big\{\underset{j\in S_k}{\mathrm{argmax}}\ y^j, \forall\ S_k\in Q\big\}\cup\big\{\underset{j\in S_k}{\mathrm{argmin}}\ y^j, \forall\ S_k\in Q\big\}$ \\
\hline
\end{tabular}
\end{small}
\end{center}
\vskip -0.1in
\end{table}

\subsection{The Choice-Based Questionnaire}
Suppose we have an unlabeled dataset of $n$ examples, $X=\{x_1, \cdots, x_n\}, x_i\in\mX$, for which we wish to obtain absolute-scale, continuous-valued labels suitable for supervised training, i.e., $Y=\{y_1, \cdots, y_n\}, y_i\in\mY\subseteq\mathbb{R}$. In choice formats, the examples are partitioned in to $n/s$ choice sets (minibatches) each of size $s$. Experts are then asked to choose the most and least risky example within each choice set, $S_k=\{x^k, \cdots, x^{k+s-1}\}, x^i\in X$. The identifier function, $t: x^{t(i)}=x_i$, is assumed to be known. The superset of choice sets encompassing all $n$ examples is called the questionnaire, $Q=\{S_1, \cdots, S_{n/s}\}$. There may be multiple questionnaires $Q_1, \cdots, Q_q$ assigned to multiple experts; however, each questionnaire consists of the same set of examples $X$, and each example $x_i$ appears only once in each questionnaire. The partitioning (or minibatching) of choice sets should, however, be different across questionnaires to increase the diversity of pairwise comparisons (i.e., $Q_i\cap Q_j=\emptyset, \forall\ i\!\neq\! j$). A method for maximizing pairwise diversity is presented in Section \ref{sec:choice_set_optimization}. We allow the experts to have some zero-mean uncertainty $\epsilon$ in their judgment.

\subsection{From Choice to Score}
\label{sec:choice2score}
In the $l$-th questionnaire, we encode the \textit{choice} $c$ of example $i$ as follows:

\begin{equation}
    c(y_i|l)=
    \begin{dcases}
        1, & \text{if } y_i=\max_{j\in S_k}y^j \\
        -1, & \text{if } y_i=\min_{j\in S_k}y^j \\
        0, & \text{otherwise}
    \end{dcases}
\end{equation}

Here, $S_k$ is the choice set which hosts $x_i$ in questionnaire $l$. We then compute the \textit{mean choice} $\bar{c}_i$ by averaging the choice $c(y_i|l)$ for example $i$ across all $q$ questionnaires:

\begin{equation}
\label{eq:encodechoice}
    \bar{c}_i(y_i)=\frac{1}{q}\sum_{l=1}^q c(y_i|l)
\end{equation}

We seek to know the functional relationship between mean choice and risk score so that we can convert the relative choice-based questionnaire results to a stand-alone, absolute-scale measure of risk, i.e., the label $y$, for supervised training. Given a label distribution $P(Y=y)=f(y)$ as a prior, the \textit{expected choice} of example $i$ can be derived as:

\begin{equation}
\label{eq:meanagg}
\begin{split}
    \braket{c(y_i)}  & = \mathbb{E}_{y_i\sim Y}\enspace c(y_i) \\
                  & = +1\times P(y_i\! =\! \max_{j\in S_k}y^j) \\
                  &\quad -1\times P(y_i\! =\! \min_{j\in S_k}y^j) \\
                  &\quad +0\times P(y_i\text{ is neither max nor min})
\end{split}
\end{equation}

Based on independence (i.e., examples are placed into choice sets without regard for the others selected), the probability that example $i$ has the largest true label within its choice set is:

\begin{equation}
  \label{eq:probmax}
\begin{split}
    P(y_i\! =\! \max_{j\in S_k}y^j) &= \prod_{j\in S_k, j\neq i}P(y_i\! >\! y^j) \\
      &= \prod_{j\in S_k, j\neq i}\left(\int_{-\infty}^{y_i}f(y^j)\:\mathrm{d}y^j\right) \\
      &= \left(\int_{-\infty}^{y_i}f(y')\:\mathrm{d}y'\right)^{s-1}
\end{split}
\end{equation}

A similar derivation can be made for the probability that example $i$ has the smallest true label. Substituting that and Eq. \ref{eq:probmax} into Eq. \ref{eq:meanagg}, the relationship between the expected choice and risk score is:

\begin{equation}
    \label{eq:choice2score}
    \braket{c(y_i)} = \left(\int_{-\infty}^{y_i}f(y')\:\mathrm{d}y'\right)^{s-1} - \left(\int_{y_i}^{\infty}f(y')\:\mathrm{d}y'\right)^{s-1}
\end{equation}

There is no analytical inverse for Eq. \ref{eq:choice2score}, but the inverse is readily estimated by numerical optimization. In the limit of a large number of questionnaires, the mean choice (which is measured) and expected choice converge:

\begin{equation}
    \label{eq:converge}
    \lim_{q\to\infty}\bar{c}(y_i) = \braket{c(y_i)}
\end{equation}

\begin{figure}[b]
\centering
\begin{subfigure}{.24\textwidth}
    \centering
    \includegraphics[width=\textwidth]{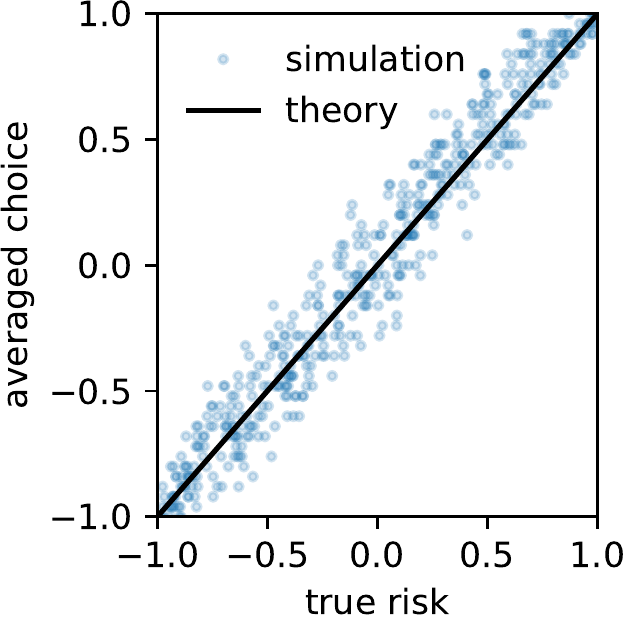}
    \caption{Choice set size = 3}
    \label{fig:setsize3}
\end{subfigure}
\begin{subfigure}{.24\textwidth}
    \centering
    \includegraphics[width=\textwidth]{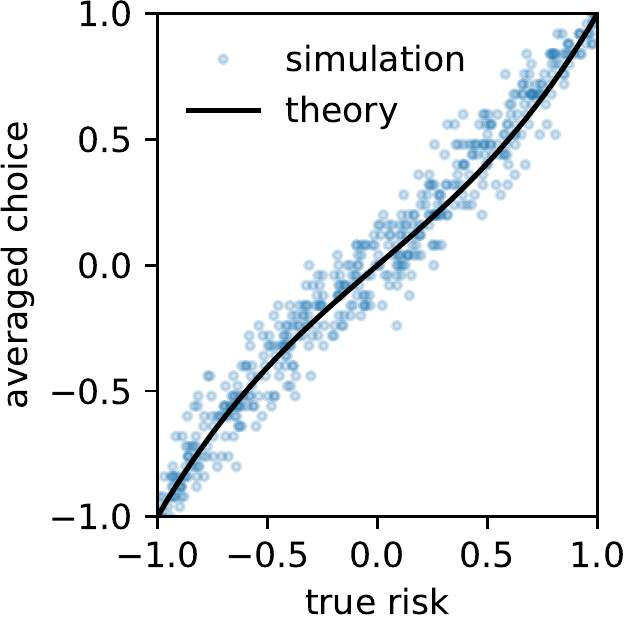}
    \caption{Choice set size = 4}
    \label{fig:setsize4}
\end{subfigure}
\begin{subfigure}{.24\textwidth}
    \centering
    \includegraphics[width=\textwidth]{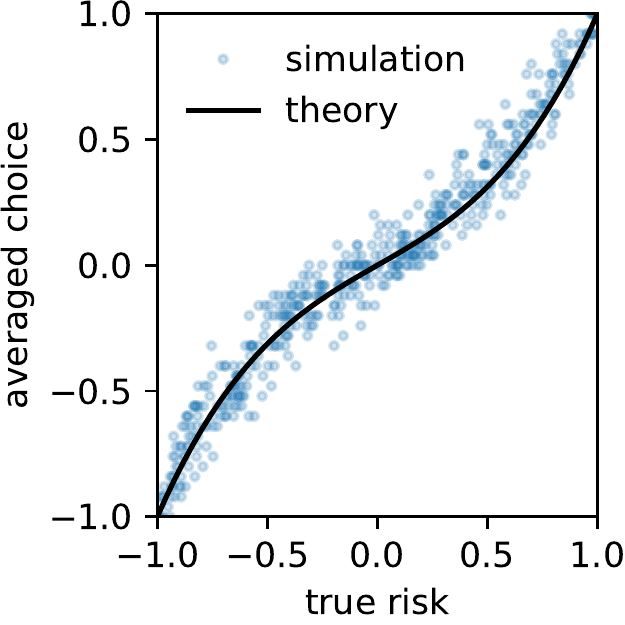}
    \caption{Choice set size = 5}
    \label{fig:setsize5}
\end{subfigure}
\begin{subfigure}{.24\textwidth}
    \centering
    \includegraphics[width=\textwidth]{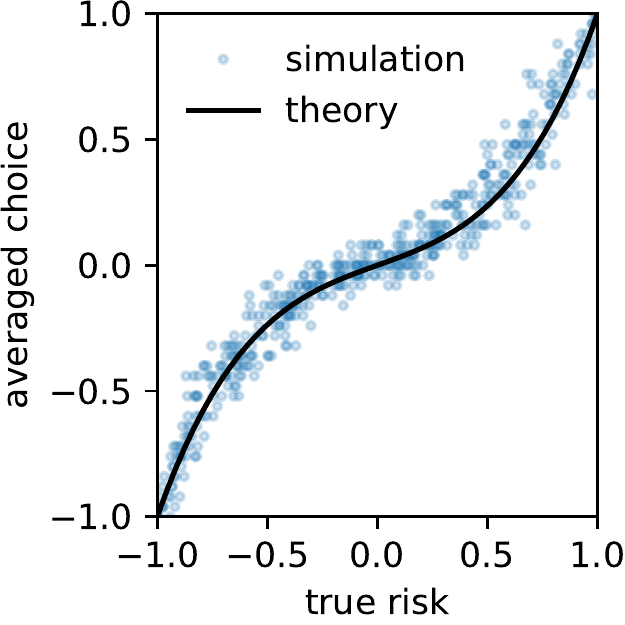}
    \caption{Choice set size = 6}
    \label{fig:setsize6}
\end{subfigure}
\caption{\small Simulation results showing averaged choice as a function of true label for 500 different examples (blue dots). An oracle evaluated each of the examples using the choice-based paradigm over 25 questionnaires. The examples' true labels are sampled from a uniform distribution. \label{fig:choice2score}}
\end{figure}

\textbf{Example} Consider a uniform label distribution $f(y)=1/2$ for $y\in [-1,1]$. The expected choice based on Eq. \ref{eq:choice2score} is:

\begin{equation}
    \braket{c(y_i)}=\left(\frac{y_i}{2}+\frac{1}{2}\right)^{s-1} - \left(-\frac{y_i}{2}+\frac{1}{2}\right)^{s-1}
\end{equation}

We conduct a simulation of 25 questionnaires each with the same set of 500 examples whose labels are sampled from a uniform distribution between -1 and 1. In each questionnaire, the examples are randomly partitioned into choice sets of size $s$. Within each choice set, an oracle is used to provide the choice $c_i$ for each example as to whether it is most/least/neither risky within its respective choice set. The choices are encoded and mean aggregated as described by Eqs. \ref{eq:encodechoice}, \ref{eq:meanagg}. Figure \ref{fig:choice2score} plots the averaged choice for each example as function of its true label. The simulation data fits our theoretical result perfectly, verifying that the choice-to-score relation in Eq. \ref{eq:choice2score} can be used to convert \textit{relative} information to \textit{absolute} information about the labels.

\section{End-to-End Algorithm for Inherent Risk Scoring}

\subsection{Overview}
We now present the Inherent Risk Model (IRM) built for a large financial institution. The data consisted of all customers active over the course of a four-year period. All entities considered were banking participants during that time period. The IRM was developed with the intended purpose of being used as a risk prioritization framework to quickly estimate the probability that a given customer's profile of Know Your Customer (KYC) variables was a high or low inherent risk for being involved in money laundering activity. The IRM's prediction of a customer's inherent risk based on KYC variables, without access to transaction information, was designed to mimic the kind of information that an anti-money laundering investigator, henceforth called Subject Matter Advisor (SMA), has access to when performing a first-level money laundering screening. Each customer (example) consisted of 24 variables (features), which included basic customer information, KYC risk flags, behavioral markers, and triggered alerts (we have redacted the specific variable names to protect the parties involved). Roughly 150,000 unique customer profiles were available.

\subsection{Synthetic Examples via Optimal Experimental Design}

\begin{figure}[b]
\vskip -.4cm
	\centering
	\begin{subfigure}[t]{.38\textwidth}
    	\centering
	\includegraphics[width=\textwidth]{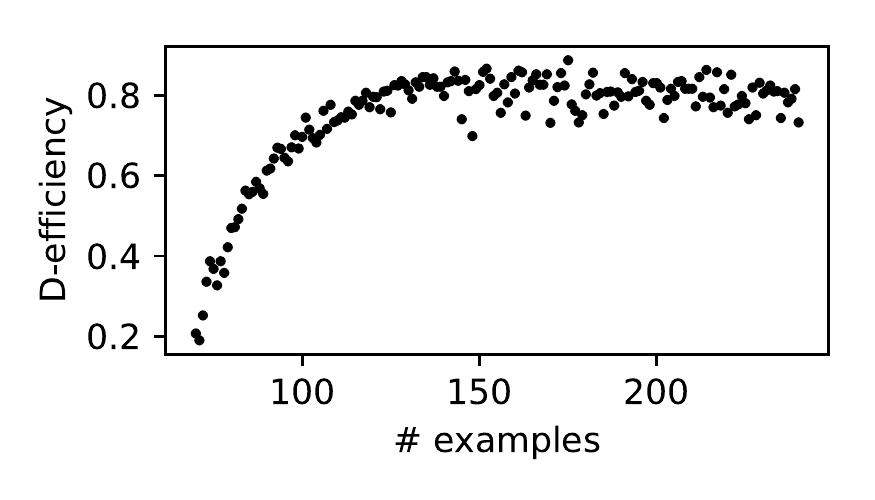}
	\caption{\small Optimal D-efficiency computed by the \code{optMonteCarlo} function in R as a function of the number of examples. There are diminishing returns after about 120 examples.
	\label{fig:deff}}
	\end{subfigure}
	\qquad
	\begin{subfigure}[t]{.38\textwidth} 
	\centering
	\includegraphics[width=\textwidth]{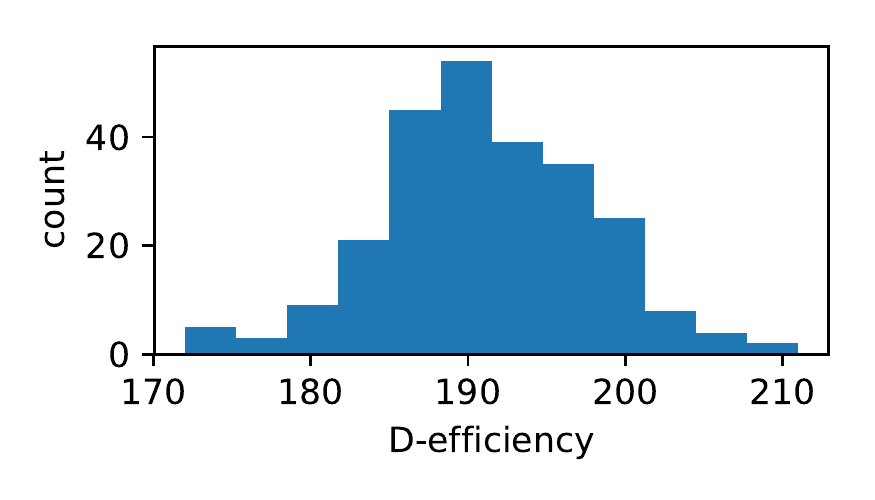} 
	\caption{\small Results of a simulation of 250 trials, for a fixed list of attributes and attribute levels, where the D-optimal number of profiles is returned from a sample size of 5 for one trial.}
	\label{fig:D_Optimal_histogram}
	\end{subfigure}
	\caption{\small Results of our optimal experimental design using the \code{optMonteCarlo} package in R. \label{fig:dopt}}
\end{figure}

While we had access to many examples (i.e., customer profiles), we did not want our trained model to depend on the particular distribution of our training data, as distributions can vary across different institutions. Plus, we did not have the SMA resources to label such a large dataset. To eliminate these issues, we created a synthetic dataset of examples using Optimal Experimental Design (OED). OED operates under the assumption that obtaining a label for each example is costly and presents a theory for selecting examples that, given labels, predict the target variable as well as possible. In particular, we seek to optimize the D-efficiency, defined as $|X^TX|$ where $X\in \mathbb{R}^{n\times m}$ is the matrix of all $n$ examples with $m$ features. Maximizing D-efficiency makes the $n$ examples span the largest possible region in feature space, ensuring that the examples are as orthogonal and balanced as possible (\cite{Eriksson2000}).

Optimal experimental design is obtained for the model discussed in this letter through the use of the Federov Coordinate Exchange algorithm via Monte Carlo implemented in the \code{optMonteCarlo} function of the R library \code{AlgDesign}. Only linear terms were considered. We observe that the D-efficiency increases with the number of examples allowed up to about 120 examples, after which the gain in D-efficiency diminishes. See Figure \ref{fig:deff}. A simulation was then performed to assess stability of this result. \code{optMonteCarlo} was called on the collection of 24 business-approved features  returning the maximum D-efficiency observed across a sample size of five, constituting one trial. Figure \ref{fig:D_Optimal_histogram} shows 250 trials and identifies an optimal example count range from 185 to 190. An example count of 188 was selected based on the group theoretic results of Section \ref{sec:choice_set_optimization}.

\subsection{Choice Set Size Selection}

\begin{figure}[b]
\vskip -.5cm
\centering
	\begin{subfigure}[t]{.32\textwidth}
    	\centering
    	\includegraphics[width=\textwidth]{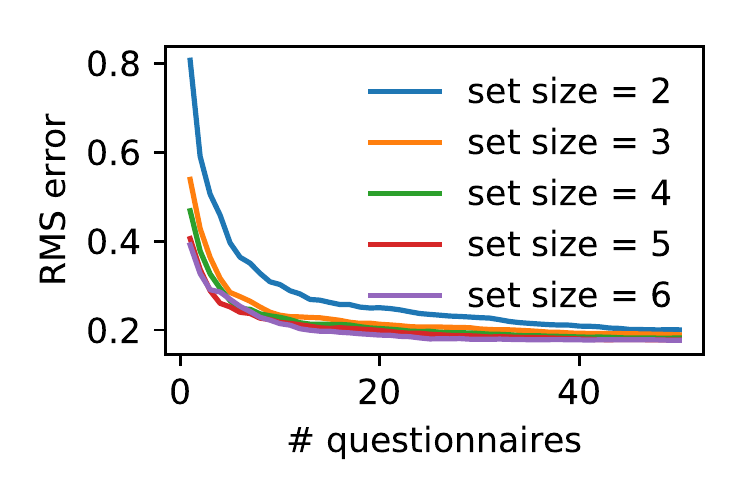}
\caption{\small RMS error between the true label and the label obtained by taking the inverse of Eq. \ref{eq:choice2score} on a set of 188 examples. As the number of questionnaires increases, the error converges to zero. A choice set size of 4 yielded the lowest error at 20 questionnaires, informing our decision of using that size. \label{fig:errorvs}}
    \end{subfigure}
    \:
	\begin{subfigure}[t]{.32\textwidth} 
    	\centering
    	\includegraphics[width=\textwidth]{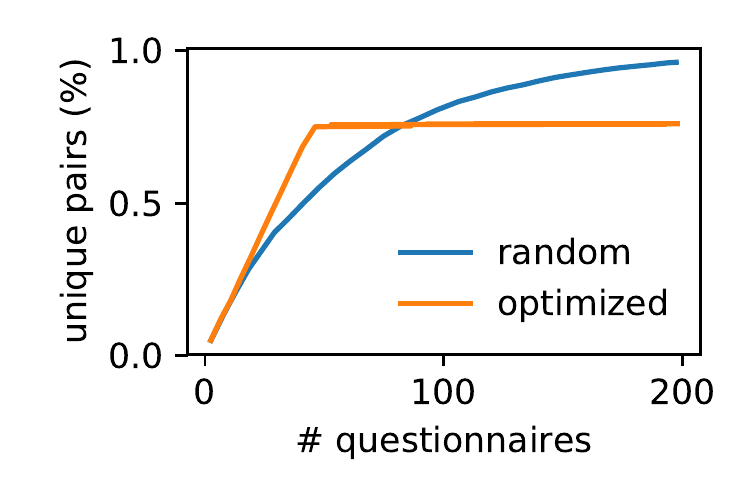} 
        \caption{\small Percentage of unique pairs within the choice sets of $q$ questionnaires, where $q$ is the x-axis. Random and optimized (group theoretic) choice sets are compared and demonstrated for 188 examples on 47 choice sets with prime number factors 13, 17, and 19. \label{fig:group_performance}}
	\end{subfigure}
	\:
	\begin{subfigure}[t]{.32\textwidth} 
    	\centering
    	\includegraphics[width=\textwidth]{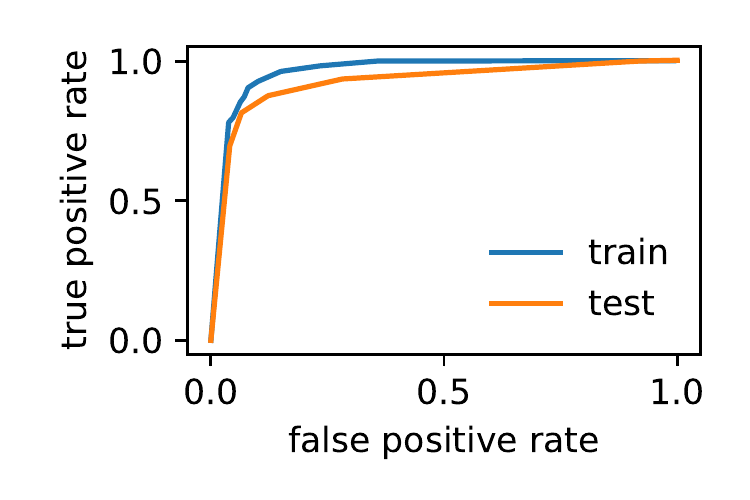} 
    	\caption{\small Inherent Risk Model ROC curve on SMA-evaluated questionnaires. The champion model was logistic regression with a threshold for positive/negative classification. Area under the curve was 97\% on train and 93\% on test.}
    	\label{fig:roc}
	\end{subfigure}
	\caption{\small Results of (a) choice set size selection, (b) choice set optimization, and (c) ROC curve of model on train and test sets.}
\end{figure}

We now determine the optimal amount of information that we can obtain from the SMAs as a function of choice set size. We attack this problem using error analysis on simulations that we performed like the one in Section \ref{sec:choice2score}. 188 profiles are sampled from a normal label distribution, and we simulate the oracle responses to $q$ questionnaires to obtain the mean choice $\bar{c}_i$. The mean choice is then converted to an estimated label by Eq. \ref{eq:choice2score}. Next, we calculate the error per example as the difference between that example's estimated label and its true label. We then take the RMS of the error across all the examples as a measure of how well the information from the SMAs approximates the ground truth. Figure \ref{fig:errorvs} shows the RMS error as a function of the number of questionnaires and choice set size. It is apparent that a choice set size of 2 has the most RMS error; this makes sense because knowing the least risky example in a 2-example choice set is redundant when the most risky example is already known, so only one pairwise comparison's worth of information is obtained. On the other hand, in choice set sizes of 3, at least 3 pairwise comparisons' worth of information is obtained between the most risky, least risky, and in-between profiles. Choice set sizes of 3 and beyond have similar RMS errors, with set sizes 4 and 5 having the minimum values. Based on this analysis, we chose a choice set size of 4.

\subsection{Choice Set (Minibatch) Optimization}
\label{sec:choice_set_optimization}
Given $q$ questionnaires, how should examples be partitioned (minibatched) into choice sets to maximize the amount of information obtained? Since the base unit of \textit{relative} information is the pairwise comparison, here we present a group theoretic method that maximizes the amount of \textit{unique} intra-choice-set pairs as a proxy to maximizing the amount of information obtained through the questionnaires. We show in Figure \ref{fig:group_performance} that our method achieves 75\% of the possible unique pairwise comparison in half the number of questionnaires as compared to random minibatching.

Choice-based analysis references do not address information-theoretic concerns about the information extracted when the ground truth is not absolutely defined. The authors are not aware of a  mechanism or technique that provides guidance on how to optimize the maximal information extracted from choice-based generated questionnaires. While optimal profiles were generated using the concept of D-optimality, the construction of the questionnaires itself is a combinatorics question handled by a computer if the problem size is small. Optimizing the arrangement of profile IDs is similar to the class of problems called backtracking. Randomly selecting the choice sets from over $10^{9}{\sim}185^{4}$ combinations and checking each against the required matching conditions that guarantee minimizing profile comparison redundancy, over a modest number of questionnaires, is not computationally tractable. We seek to develop a questionnaire generation optimization that uses each profile once per questionnaire and minimizes the redundant inferred choice set size two comparisons to approach an information-theoretic limit for information extraction. We achieve a general solution to this problem by constructing a reducible matrix representation of a finite group whose cycle corresponds to the maximum number of unique questionnaires that can be constructed given the representation.

We can motivate this approach by discussing the pairwise combinations of choice set profiles found in a choice set of size 4. The SMA’s act of observation is on one choice set at a time. The SMA assesses the relative risk of the profiles presented in that choice set, selecting one profile as highest relative risk and one profile as lowest relative risk. From one SMA's evaluation of a choice set of size four, we can infer two of the four size three inferred choice set evaluations, and five of the six size two choice set evaluations. On four elements, five of the six pairs are implied. It is the opinion of the authors that this comparison of different choice sets sizes is similar to an analysis of reducible finite group representations as they are compositions of irreducible representations. We use this insight to generate a group theoretic questionnaire optimization of choice sets. The uniqueness of the result guarantees that any other sampling methodology obtaining this rate of unique pairs per questionnaire is isomorphic to the underlying group representation we construct. This also implies that no other algorithmic process can obtain a faster unique enumeration of pairs for a collection of choice sets.

We present this result as an application of the Polya Enumeration Theorem in Algorithm \ref{alg:generate_group_rep}.

\vskip -.1cm
\begin{algorithm}[H]
  \caption{Choice Set Diversifier - Size 4}
  \label{alg:generate_group_rep}
\begin{algorithmic}

  \STATE \textbf{Input}: Range of D-optimal profiles $R$
  \STATE \textbf{Select} $c\in\mathbb{N}$ and $c\in R$ such that $c=4 p$ where $p$ is prime; $p$ is the number of choice sets
  \STATE \textbf{Select} three different prime numbers $p_{1},p_{2},p_{3}$ such that $3 <p_{i}$ and $p<p_{i}p_{j}$ for $ i\neq j$.
  \STATE Without replacement, randomly \textbf{Assign} each of the $4 p$ profiles a profile key from $1$ to $4 p$ and place them into equal sized lists: $[A, B, C, D]$
  \STATE \textbf{Construct} a $rank(p)$ square identity matrix, $\mathbb{I}$
  \STATE \textbf{For each} prime number $p_{1},p_{2},p_{3}$:\\
    \setlength{\leftskip}{0.5cm} \textbf{Construct} a $rank(p)$ square permutation matrix, $\hat{G}_{p_{1}}$, $\hat{G}_{p_{2}}$, $\hat{G}_{p_{3}}$:\\
    Where elements of $[B, C, D]$ are mapped by position index $i$ in each list such that\\
    $i\mapsto i + p_{1}$ modulo $p$ \textbf{OR} $i\mapsto i + p_{2}$ modulo $p$ \textbf{OR} $i\mapsto i + p_{3}$ modulo $p$

    \setlength{\leftskip}{0cm}
    \STATE \textbf{Define a Group Action $\hat{\textbf{G}}$} on the vector space of ordered tuples $\braket{a,b,c,d}$ where \\
    \setlength{\leftskip}{0.5cm}$a\in A, b\in B, c\in C, d\in D$ such that $\hat{G}=\mathbb{I}\oplus\hat{G}_{p_{1}}\oplus\hat{G}_{p_{2}}\oplus\hat{G}_{p_{3}}$ and $\hat{G}^{(p+1)}=\hat{G}$

    \setlength{\leftskip}{0cm}
    \STATE \textbf{Action on a Choice Set} is given by\\
    \setlength{\leftskip}{0.5cm}
    $\hat{G}(\braket{a_{1},b_{1},c_{1},d_{1}})=\mathbb{I}*a_{1}\oplus\hat{G}_{p_{1}}*b_{1}\oplus\hat{G}_{p_{2}}*c_{1}\oplus\hat{G}_{p_{3}}*d_{1}=\braket{a_{1}, b_{1+p_{1}}, c_{1+p_{2}}, d_{1+p_{3}}}$
    
    \setlength{\leftskip}{0cm}
    \STATE \textbf{Append} each ordered list $[A,B, C, D]$ so that $U= A+B+C+D$. $U$ is the first questionnaire
    \STATE \textbf{Apply $\hat{\textbf{G}}$} to the randomized ordered list $U$ to generate the next questionnaire
    \STATE \textbf{Stop} when $\hat{G}$ has been applied to $U$ $p$ number of times
    \STATE \textbf{Return} $p$ number of unique questionnaires $U$, $U'$, $U''$, $\cdots$
\end{algorithmic}
\end{algorithm}
\vskip -.3cm

We now have four permutation matrices (one is the identity matrix) that separately act on a choice set element. If the list of choice sets is a $4*p$ length row vector comprising one questionnaire, $\hat{G}$ becomes a $4*p$ dimensional representation with subsequent applications of $\hat{G}$ generating unique questionnaires until repeating on the $p+1$ questionnaire. This generates an optimal collection of questionnaires where each profile is used once per questionnaire, each choice set combination appears at most once for all questionnaires, and any two out of the four profiles in any choice set appear at most once over all questionnaires. This result motivated our choice of $n=188$ examples with prime number factors 13, 17, 19.

\subsection{Training and Evaluation}

\textbf{Synthetic Data.} Our inherent risk scoring model was trained, validated, and tested on the collected questionnaire results from a team of SMAs. The questionnaires consisted of unique choice sets of synthetic KYC profiles as per the complete methodology outlined in this paper. We trained a linear classifier on the 188 customer profiles from the training set. We separately obtained a test set of 52 examples from the same SMAs via the same approach. We define a threshold in the range of [-1,1] to separate profiles that are considered high (e.g., positive prediction) and low risk (e.g., negative prediction). The ROC curve as a function of this threshold is shown in Figure \ref{fig:roc}. We tuned the threshold to be compliant with an industry standard false-negative (misclassifying a high risk customer as low risk) rate of roughly 1:1,000. At this threshold, we achieved an area under the curve (AUC) of 97\% on train and 93\% on test, with a classification error of 8\% on train and 11\% on test.

\textbf{Real-World Data.} The same IRM model was applied to the customer population and provided a rank-ordering of those profiles. A logic layer stating the presence of 1 of 14 scenarios associated with risky trading, transaction, or settlement behavior conditioned the sample down to roughly 8,000 unique customers. The remaining alerted accounts were individually reviewed, providing a rare performance evaluation for the model on real-world data. The prioritization framework created with the authors' complete methodology was tested by letting it select the 1,500 profiles it predicted to be the riskiest, partitioning the alerting population. After SMAs reviewed the 4,000 profiles and identified which ones should be recommended for escalation, the IRM capability to pick truly ``riskier'' profiles resulted in a 15.5x improvement in the escalation rate. Compared with the escalation rate associated with a random alerting profile of 0.775\%, the IRM achieved a 2.4x improvement. None of the 31 SMA-recommended escalations were in the bottom 40\% of the IRM risk score, highlighting a low false-negative rate.

\begin{table}
\vskip -.2cm
\caption{IRM performance on SMA-evaluated questionnaires.}
\label{tab:perf}
\begin{center}
\begin{small}

    \begin{tabular}{llll}
    \hline
    Metric & Train & Test \\ 
    \hline
    AUC & 97\% & 93\% \\ 
    Classification Error & 8\% & 11\% \\
    \hline
    \end{tabular}
    \vskip 0.05in

    \begin{tabular}{llll}
    \hline
    Population Group & Profiles & SMA Escalations & Escalation Rate\\
    \hline
    IRM Selected Alerted Profiles & 1,500 & 28 & 1.87 \\
    Remaining Scenario Alerted Profiles & 2,500 & 3 & 0.12 \\
    \hline
    \end{tabular}

\end{small}
\end{center}
\vskip -0.1in
\end{table}

\section{Related work}

\cite{Ratner2016} presented the idea of programmatically generating (possibly noisy) training data using distant supervision and rule-based labeling functions. The fact that noisy, crowdsourced training data may still be used to achieve high accuracy models forms the basis of our approach. Generating labels from alternative feedback (other than direct labels) is not new: much work has been done in the field of learning to rank via pairwise preferences (\cite{Hull2008, Jamieson2011}). \cite{Xu2018} considers the leveraging of ranking or comparison information as a way to improve on their model, but their method assumes that some number of direct labels are known, whereas we do not make this assumption. Choice-based conjoint analysis has been widely used in preference learning (\cite{Asioli2016}) in marketing, but has not been applied to risk assessment.

\section{Conclusion}

We have introduced a novel end-to-end methodology for developing machine learning models trained on labels that are defined by relative comparisons in the absence of absolute label definitions. Our model, used for inherent risk prediction, was trained on a set of choice-based responses to group theoretic optimized questionnaires consisting of synthetic examples. The model was applied to real-world profiles to evaluate the money laundering risk of banking customers. We achieved a 15.5x improvement of the identification rate of customer profiles that money laundering experts recommended for escalation. Broadly, these results challenge the community to consider the class of machine learning problems where expert feedback is missing absolute label definitions and labels must be gathered from alternative feedback methods, such as choice.

Two shortcomings of our approach are listed here to stimulate further research. First, our formulation assumes oracle questionnaire respondents who choose the most/least risky customer in each choice set with no mistakes with regard to their ground truth risk score. In reality, human experts in this domain will have differing biases and noise in their responses. Second, we assume that the label distribution is known. While it is reasonable to assume a normal distribution, based on the central limit theorem, in the case of a linear model and D-optimal example set, this is a restrictive condition.

\subsubsection*{Acknowledgments}
The people referenced in this acknowledgements section are the authors' colleagues at Ernst \& Young LLP, and we thank them for their support. The authors thank Jonathan DeGange and Carl Case for detailed preliminary technical conversations that seeded this work. Jonathan DeGange's contributions as an expert in Monte Carlo methods were critical to this work. Nick Brennan, Zachary Carideo, and Maoxin Ye provided data science support and experimentation. Finally, the authors thank Ron Giammarco, Darrin Williams, Sameer Gupta, Carl Case, Ali Khan, Jonathan DeGange, Joe Kruse, Greg Capece, and Rajarajan Sampath for project oversight and support. The results discussed in this letter and references to terms {\em accurate}, {\em efficient}, and {\em bias} are with respect to the letter’s mathematical treatment of a generalized methodology framework. The views and conclusions expressed in this material are those of the authors and should not be interpreted as representing the official policies or endorsements, either expressed or implied, of Ernst \& Young LLP.

\bibliography{nips_2018}

\begin{thebibliography}{7}
\providecommand{\natexlab}[1]{#1}
\providecommand{\url}[1]{\texttt{#1}}
\expandafter\ifx\csname urlstyle\endcsname\relax
  \providecommand{\doi}[1]{doi: #1}\else
  \providecommand{\doi}{doi: \begingroup \urlstyle{rm}\Url}\fi

\bibitem[Asioli et~al.(2016)Asioli, Nass, Avrum, and Almli]{Asioli2016}
D.~Asioli, T.~Nass, A.~Avrum, and V.L. Almli.
\newblock Comparison of rating-based and choice-based conjoint analysis models.
  a case study based on preferences for iced coffee in norway.
\newblock \emph{Food Quality and Preference}, 48:\penalty0 174 -- 184, 2016.
\newblock URL
  \url{http://www.sciencedirect.com/science/article/pii/S0950329315002499}.

\bibitem[Sethuraman et~al.(2005)Sethuraman, Kerin, and Cron]{sethuraman2005}
Raj Sethuraman, Roger~A Kerin, and William~L Cron.
\newblock {A field study comparing online and offline data collection methods
  for identifying product attribute preferences using conjoint analysis}.
\newblock \emph{J. Bus. Res.}, 58\penalty0 (5):\penalty0 602--610, 2005.
\newblock ISSN 0148-2963.
\newblock \doi{https://doi.org/10.1016/j.jbusres.2003.09.009}.
\newblock URL
  \url{http://www.sciencedirect.com/science/article/pii/S0148296303002029}.

\bibitem[Eriksson et~al.(2000)Eriksson, Johansson, Kettaneh-Wold, Wikstrom, and
  Wold]{Eriksson2000}
L.~Eriksson, E.~Johansson, N.~Kettaneh-Wold, C.~Wikstrom, and S.~Wold.
\newblock \emph{Design of Experiments: Principles and Applications}.
\newblock Learnways AB, 2000.

\bibitem[Ratner et~al.(2016)Ratner, De~Sa, Wu, Selsam, and R\'{e}]{Ratner2016}
Alexander~J Ratner, Christopher~M De~Sa, Sen Wu, Daniel Selsam, and Christopher
  R\'{e}.
\newblock Data programming: Creating large training sets, quickly.
\newblock In \emph{Advances in Neural Information Processing Systems 29}, pages
  3567--3575. 2016.
\newblock URL
  \url{http://papers.nips.cc/paper/6523-data-programming-creating-large-training-sets-quickly.pdf}.

\bibitem[Hullermeier et~al.(2008)Hullermeier, Farnkranz, Cheng, and
  Brinker]{Hull2008}
Eyke Hullermeier, Johannes Farnkranz, Weiwei Cheng, and Klaus Brinker.
\newblock Label ranking by learning pairwise preferences.
\newblock \emph{Artificial Intelligence}, 172\penalty0 (16), 2008.
\newblock URL
  \url{http://www.sciencedirect.com/science/article/pii/S000437020800101X}.

\bibitem[Jamieson and Nowak(2011)]{Jamieson2011}
Kevin~G. Jamieson and Robert~D. Nowak.
\newblock Active ranking using pairwise comparisons.
\newblock In \emph{Advances in Neural Information Processing Systems}. 2011.

\bibitem[Xu et~al.(2018)Xu, Muthakana, Balakrishnan, Singh, and
  Dubrawski]{Xu2018}
Yichong Xu, Hariank Muthakana, Sivaraman Balakrishnan, Aarti Singh, and Artur
  Dubrawski.
\newblock Nonparametric regression with comparisons: Escaping the curse of
  dimensionality with ordinal information.
\newblock In \emph{Proceedings of the 35th International Conference on Machine
  Learning, {ICML} 2018, Stockholmsm{\"{a}}ssan, Stockholm, Sweden, July 10-15,
  2018}, 2018.
\newblock URL \url{http://proceedings.mlr.press/v80/xu18e.html}.

\end{thebibliography}
\end{document}



\appendix

\section{Illustrations of Poisoning and how the decision boundary will rotate after training with the poison instances}
\ref{fig:schem} illustrates our poisoning scheme. When a network is trained on clean data + the very few poison instance, the linear decision boundary in feature space (Figure \ref{fig:featspace}) is expected to rotate to include the poison instance in the base class side of the decision boundary in order to avoid misclassification of that instance.

\begin{figure}[h]
\centering
\begin{subfigure}{.55\textwidth}
    \centering
    \includegraphics[width=1\textwidth]{fig_schem}
    \caption{ }
    \label{fig:schem}
\end{subfigure}%
\begin{subfigure}{.45\textwidth}
    \centering
    \includegraphics[width=1\textwidth]{fig_featspace}
    \caption{Illustration of the feature space (activations of the penultimate layer before the softmax layer of the network) representation of clean training data, poison instance, and target instance. Note that the target instance is \textit{not} in the training set. Thus it will not affect the loss when the nearby poison instance causes the decision boundary to shift to encompass both of them into the base region.}
    \label{fig:featspace}
\end{subfigure}
\caption{(a) Schematic of the clean-label poisoning attack. (b) Schematic of how a successful attack might work by shifting the decision boundary.}
\label{fig:introPics}
\end{figure}

\section{Comparison to adversarial examples}
The clean-label targeted poison attack is similar to adversarial examples in the sense that they both are used for misclassifying a particular target instance. 
However, they differ in the kinds of freedom the attacker has on manipulating the target.
Adversarial examples assume that the target can be slightly modified and hence they craft an example which looks very similar to the target instance in input space but gets misclassified. In the targeted clean-label framework, we assume that the attacker has no control over the target instance during test time and can not (or is not willing to) modify it even slightly. 
This makes the threat posed more concerning, as it allows one to control the classifier's decisions on test-time instances outside their spectrum of control.
This framework could be also useful for fooling a face recognition system. 
While it has been shown that an adversary can craft accessories for a target such that wearing that accessory causes the face-recognition system to misclassify the target, there are many sensitive situations which the target is prevented from wearing any accessories.

\section{Sampling the candidate target instance for more success}
As mentioned in the main body, a smart attacker will maximize her chances of success by choosing an effective and easy to manipulate target instance. Because outliers are separated from most training samples, the decision boundary can easily change in this location without substantially affecting classification accuracy. Also, points chosen near the decision boundary require less manipulation of the classifier in order to change its behavior. Note that the 2D illustration in Figure \ref{fig:featspace} belies the relative ease with which this condition can be fulfilled in higher dimensional spaces; nonetheless, this condition provides a heuristic by which we chose our target instances when we want higher success rates.

\section{Network architecture for CIFAR-10 classifier}
The scaled down AlexNet architecture attacked during the end-to-end experiments is summarized in Table~\ref{tab:alexNet}. Without poisoning, the network has a training accuracy of 100\% and a test accuracy of 74.5\%. To reach this accuracy, the network is trained for 200 epochs on clean data (non-poisoned) with a learning rate that decays with a schedule. The final learning rate is the one used for retraining the model once the training data set is poisoned.



\FloatBarrier
\begin{table}[H]
    \centering
    \vspace{-3mm}
    \caption{Network architecture for the poisoning of CIFAR-10 experiments.}
    \label{tab:alexNet}
    \begin{tabular}{|c|c|c|c|}
        \hline
         &Type & Kernel Size & \#Out Dim.\\
         \hline
         1&Conv+ReLU&$5\times5$&64\\
         2&MaxPool $1/2$&$3\times3$&64\\
         3&LRN&-&64\\
         \hline
         4&Conv+ReLU&$5\times5$&64\\
         5&MaxPool $1/2$&$3\times3$&64\\
         6&LRN&-&64\\
         \hline
         7&FullyConnected+ReLU&-&384\\
         \hline
         8&FullyConnected+ReLU&-&192\\
         \hline
         9&FullyConnected&-&10\\
         \hline
    \end{tabular}
\end{table}

\section{What do watermarked poisons look like?}
The poison instances' appearance depends on the target instance being attacked and also the level of transparency (opacity) of the target instance being added to the base instances for making poison instances (Fig.~\ref{fig:poisonImgs}). Attacking target instances from the birds class using poison instances built from base instances belonging to the dog class (Fig.~\ref{fig:dog_bird_30}) are both more successful and the poison instance images look less disturbed than the airplane-vs-frog attack (Fig.~\ref{fig:frog_plane_30}). For example, the complete set of 60 poison examples used for successfully attacking a target bird is illustrated in Fig.~\ref{fig:allpoisons4Test980}). If the auditor does not know the target instance (similar to the situation in Fig.~\ref{fig:allpoisons4Test980}) and is not seeing all of the poison instances side-by-side, the chances that the poison instances would be flagged as threats should not be very high. 
\begin{figure}
    \centering
    \includegraphics[width=\textwidth]{test980_60Poisons_30opac.pdf}
    \caption{\textit{All} 60 poison instances that successfully cause a bird target instance get misclassified as a dog in the end-to-end training scenario. An adversarial watermark (opacity 30\%) of the target bird instance is applied to the base instances used for making the poisons.}
    \label{fig:allpoisons4Test980}
\end{figure}

When multiple objects are present in am image, they could be exploited to craft an attack. For example, many images of the ImageNet data set contain multiple objects from different classes, although the image has only one label. A clever watermarking for these situations could add the target instance or parts of the target instance with a higher opacity in a place where it seems innocuous. For example, one can add the airplane target instance flying in the background of a poisoned frog instance. Or if the task is to misclassify Bob as Alice, the attacker can add images of Bob to group photos in which Alice is present. 

\begin{figure}[h]
\begin{center}
\centerline{\includegraphics[width=0.8\linewidth]{sampleImgs2.png}}
\caption{Poison frogs.  Every row contains optimized poisoning instances built from random images belonging to the base class (frog) for the given candidate target instance that belongs to the airplane class. We apply an adversarial watermark (a transparent overlay of the target instance ``airplane'' image) with different opacity levels. These poisoning instances are close to the airplane in feature space. For CIFAR-10 images, 30\% opacity watermarks are often hard to recognize and an unassuming human labeler would almost certainly properly label these images as ``frog.'' However, for opacity 50\% the watermark is noticeable. 
}
\label{fig:poisonImgs}
\end{center}
\end{figure}

\begin{figure}
    \centering
    \includegraphics[width=\linewidth]{frog_plane-min.png}
    \caption{Some samples of poisons made for different plane targets using base instances as frogs. Note that, the bases were watermarked with a 30\% opacity of the target. The target instance is on the top row and some of the poisons are below it. A green dot is used when the attack was successful and red dots indicate unsuccessful attacks. Note that the images do not always look clean compared to the bird vs dog task presented in Fig.~\ref{fig:dog_bird_30}. But it is important to note that when the target is not known, it may be hard to recognize it's existence in all images.}
    \label{fig:frog_plane_30}
\end{figure}

\begin{figure}
    \centering
    \includegraphics{dog_bird.png}
    \caption{Some sample target and poison instances for the task of attacking a target instance from the bird class using base instances from the dog class. The columns with the green dot above them are successful attacks. Similar, to the other experiment (attack), a 30\% opacity of the target is added to the base instance. Note that the poison instances here are better looking than the poison instances of the plane-frog task. Also, the attacks are more successful for this task. To verify that this attack is hardly detectable to an non-suspicious label auditor, one should only look at one column and ignore the top row because the auditor/labeler is not aware of the target.}
    \label{fig:dog_bird_30}
\end{figure}

\subsection{Ablation study:  How many frogs does it take to poison a network?}
We perform a leave-one-out ablation study to show the effect of each of the poisoning methods described above. Feature representations for attacking a particular target instance (before and after adversarial training) are visualized in Fig.~\ref{fig:ablation}. Results are shown for a process using all of the methods described above (row a), which produces a successful attack.
 
Row (b) shows a process that only uses one base to produce all the poisons. Training with multiple adversarial instances from the same base is called ``adversarial training,'' and makes the network robust against adversarial examples \citep{goodfellow2014explaining}.   For this reason, reusing the base image causes poisoning to fail. 
Row (c) shows a process that excludes optimization to produce feature collisions, and (d) shows a process that leaves out watermarking.

\begin{figure}
    \centering
    \begin{subfigure}{\textwidth}
        \centering
        \includegraphics[width=\linewidth]{abl_base-min.pdf}
        \caption{Successful attack - includes: optimization + watermarking + multi-base}
        \label{fig:abl_base}
    \end{subfigure}
    \begin{subfigure}{\textwidth}
        \includegraphics[width=\linewidth]{abl_1-min.pdf}
        \caption{Unsuccessful attack - includes: optimization + watermarking ; excludes: multi-base}
        \label{fig:abl_1}
    \end{subfigure}
    \begin{subfigure}{\textwidth}
        \includegraphics[width=\linewidth]{abl_2-min.pdf}
        \caption{Unsuccessful attack - includes: multi-base + watermarking ; excludes: optimization}
        \label{fig:abl_2}
    \end{subfigure}
    \begin{subfigure}{\textwidth}
        \includegraphics[width=\linewidth]{abl_3-min.pdf}
        \caption{Unsuccessful attack - includes: optimization + multi-base ; excludes: watermarking}
        \label{fig:abl_3}
    \end{subfigure}
    \caption{Ablation study showing the effect of different components of the poisoning attack on a network with end-to-end training.  The target is in 
    the ``airplane'' class, and has roughly 70\% class probability of belonging to its home class.
    Dark green dots are bases, red squares are poisons, and the dark blue triangle is the target.  Translucent points are other data points in each class.   
    (a) Two epochs of training on a successful attack that includes watermarking, 50 poison frogs from random bases, and optimization. The base instances are skewed towards the target image because they have 30\% opacity of the target. It can be seen that most of the poisoning happens during the first epoch of retraining. 
    The other rows depict unsuccessful attacks. (b) Multiple poisons are used, but all from the same base.  (c) Watermarking and multi-base poisons are used, but without optimization to collide the feature representations of the poisons with the target. Unlike the other rows, the base instances (dark green) do not include the watermarking while the poisons do. (d) No watermarking.}
    \label{fig:ablation}
\end{figure}


\maketitle
\begin{abstract}
  Inherent risk scoring is an important function in anti-money laundering, used for determining the riskiness of an individual during onboarding before fraudulent transactions occur. It is, however, often fraught with two challenges: (1) inconsistent notions of what constitutes as high or low risk by experts and (2) the lack of labeled data. This paper explores a new paradigm of data labeling and data collection for risk prediction. The data labeling is choice-based; the expert does not provide an absolute risk score but chooses the most and least risky example out of a small choice set, allowing experts to make only relative judgements of risk. The data collection is synthetic; examples are crafted using experimental design methods, obviating the need for real data which is often difficult to obtain (due to regulation or privacy concerns). We present the methodology of an end-to-end risk scoring system that we built for a large financial institution. The system was trained on a small set of synthetic data ($188$ examples) whose labels are obtained via the choice-based paradigm using a small budget of expert labelers. The system achieves 89\% accuracy on a test set of \ronny{X} examples, with a F1 score of \ronny{FILL}.
\end{abstract}

\ronny{Questins for miguel:
Can you add some content in for Section 3.1 on feature selection? I'm not exactly sure how the features (aka attributes) were selected from the many that were available.
}

\section{Introduction}

Financial institutions today are tasked with Know Your Customer obligations in order to mitigate money laundering activity in their systems that often serves to enable terrorism or drug trafficking. Banks as a result have collected large amounts of background information about their customers, such as their source of funds, business operations, and financial statements. Accurately predicting customers' {\em inherent risk} from this information is a critical function in preventing money laundering and other fraudulent behaviors {\em before} they happen.

Despite the abundance of customer information (features), there is a paucity of data of actual instances of fraud (labels). Traditionally, expert fraud investigators are employed to sift through large amounts of customer records and provide their judgments on (i.e., act as a prediction module for) the risk level of each individual. Because such a process is costly, manual, and time-consuming, much work as focusing on building accurate machine learning models for predicting inherent risk score.

Inherent risk is a concept that most find easy to evaluate in {\em comparisons}. For example, one might easily judge that a politically exposed person with frequent transactions over \$10k to foreign parties is inherently riskier {\em compared to} a customer with smaller, more stable transactions, who in turn is inherently riskier compared to a long-established domestic customer on a fixed-term deposit. However, due to the subjective and ever-changing nature of the notion of risk, it is far more difficult to judge the inherent risk of an individual on an absolute {\em quantitative} scale, e.g., between -1 and 1, than it is to make comparisons between different examples. In practice, fraud investigators are often asked only to provide labels on a binary scale of either ``high'' or ``low'' risk. While such binning strategies may make it easier for investigators, they also allow for highly inconsistent and noisy labels, because different investigators have different ideas of what constitutes as, e.g., ``high'' risk.

\subsection{Contributions}

To address the noisy and subjective nature of inherent risk labeling, we present a novel method for obtaining absolute-scale, continuous-valued labels of inherent risk from purely choice-based answers to a questionnaire from a crowd of expert labelers. Contributions include the following. (1) Leveraging Monte-Carlo D-optimal design-of-experiments approach for generating set of synthetic customer examples which covers the input space without bias. (2) An optimal algorithm for generating choice sets which minimize redundant pairings. (3) An algorithm which aggregates choice-based questionnaire responses into continuous target values with maximal, unbiased use of the information provided. (4) An outlier removal method to remove the influence of ``outlier'' labelers in a crowdsourced setting. Finally, we show that our end-to-end algorithm learned from such labels achieves a F1 score of X on a test set of customers at a large financial institution. This is, to the authors' knowledge, the first time choice-based conjoint analysis techniques have been successfully applied to financial fraud detection. 

\begin{figure}[!t]
\centering
\begin{subfigure}{.4\textwidth}
    \centering
    \includegraphics[width=\textwidth]{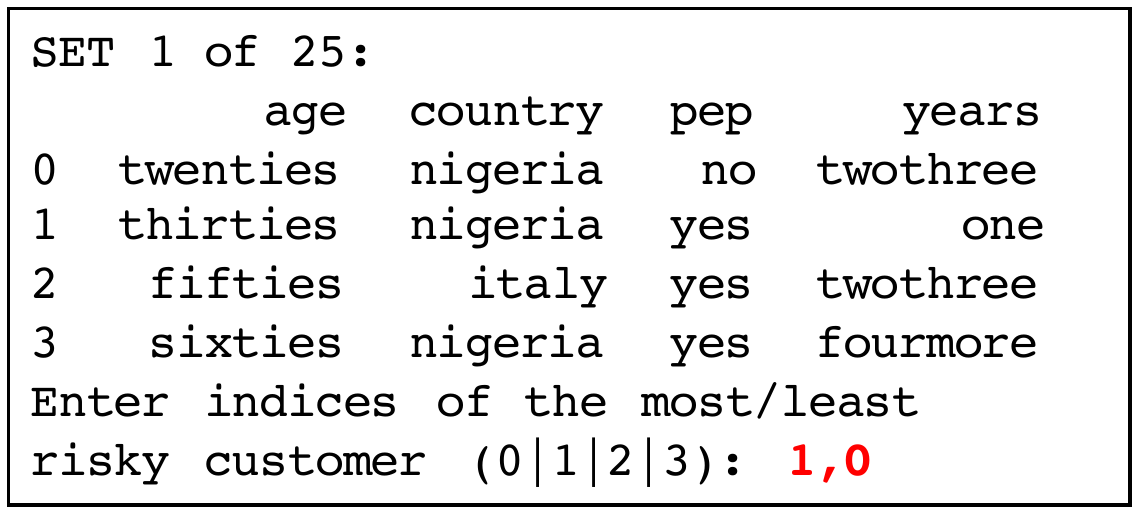}
    \caption{Example of a choice set.}
    \label{fig:question}
\end{subfigure}
\begin{subfigure}{.4\textwidth}
    \centering
    \includegraphics[width=\textwidth]{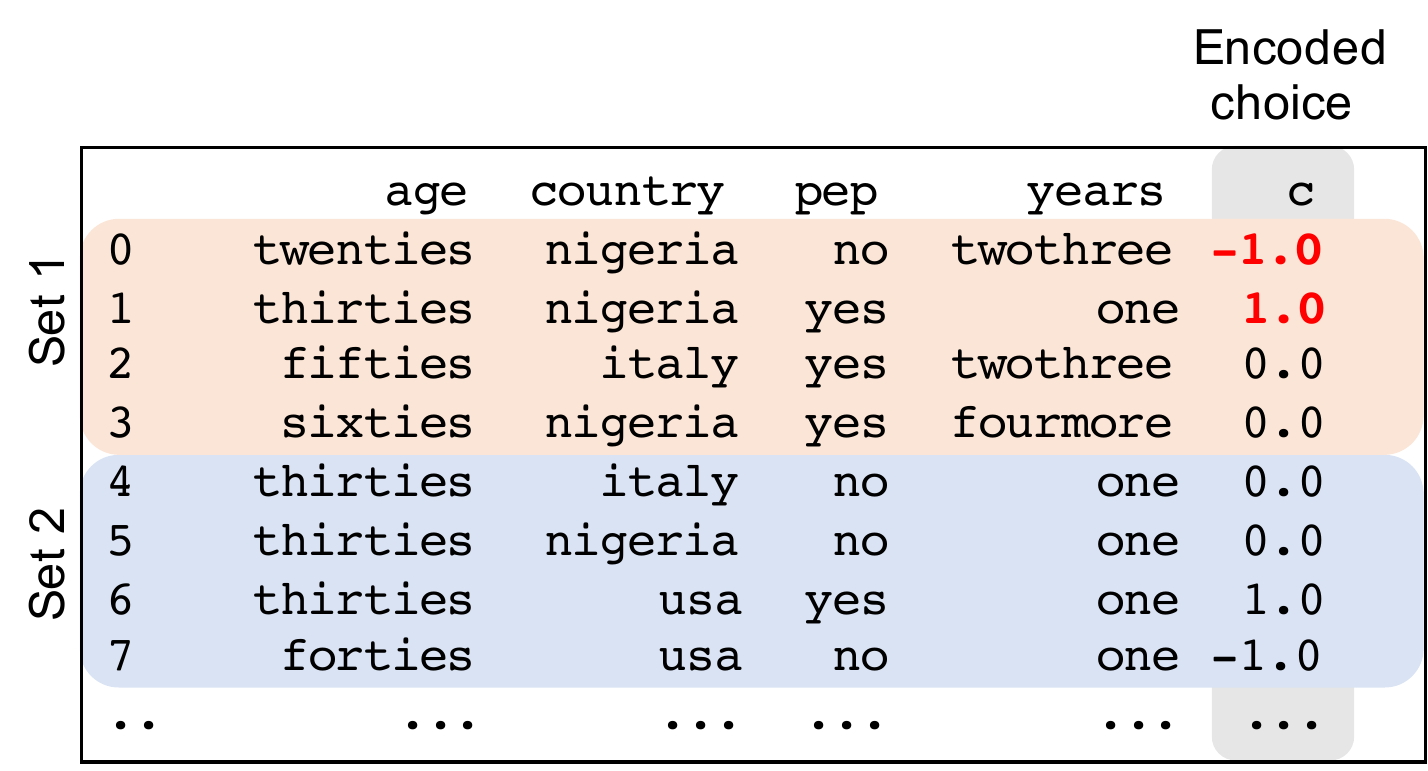}
    \caption{Example data layout.}
    \label{fig:datalayout}
\end{subfigure}
\caption{\small Illustration of the choice-based format for querying the labelers. (a) Labelers are presented with a small minibatch (choice set size of 4 here) of examples and they are asked only to select only the highest and lowest risk example from within that set. (b) Layout of the data obtained after labeling of a choice-based questionnaire. Most/least risky examples from each choice set are encoded as 1/-1 while unselected examples are encoded as 0. These encodings are mean-aggreggated over multiple questionnaires (with different minibatching) to obtain the final label. See Section \ref{sec:respagg} for more discussion.}
\label{fig:choiceexample}

\end{figure}

\section{The Choice-Based Labeling Paradigm}
We first motivate the need for choice-based label querying before diving into the analysis.

\subsection{Conjoint Analysis: A Survey of Label Querying Formats}
Conjoint analysis is a method wherein the responses from one or more labelers (who have access to the oracle labeling function with some error, i.e. $y=f(x)+\epsilon$) are used to label a set of examples for supervised training. The labels can be queried in several formats, including direct, ranking, or choice formats. Table \ref{tab:conjoint} summarizes their properties.

The direct label format is typically assumed in most machine learning formulations (i.e. $y^{[i]}$ is known for every example $x^{[i]}$). However, in risk scoring, the labeler often has a relative and noisy notion of the true label and is unable to provide an accurate $y^{[i]}$. Binning the label value into discrete ordinal values (e.g., high/medium/low) is one solution but still suffers from subjectivity.

In contrast, the ranking and choice formats only require the labeler to make pairwise comparisons which alleviate the need for an absolute scale. Such labeling formats are often employed in marketing to learn customer preferences on variations of potential products \ref{todo}, where again the labeler (customer) has only a relative notion of the true label (utility of a product).

The ranking format asks the labeler to sort the entire set of examples according to their risk score, a task which can be broken down into a series of pairwise comparisons. However, it is time-consuming for a human labeler to sort hundreds or thousands of individuals---using MergeSort, one could achieve $O(n\log n)$ time complexity at best. In the choice format, the labeler is presented with small choice sets, or minibatches, of examples and asked only to select the most and least risky examples from within that set. In addition to having a time complexity of $O(n)$, choice based formats tend to be most natural for humans labelers \cite{sethuraman2005} because human labelers can ``eye-ball'' the choice set and quickly determine the examples at the extremes of the risky/non-risky spectrum, without needing to deliberate between the relative risks of those examples in between. For these reasons, the choice-based format is most popular in conjoint analysis and is the format that we will henceforth consider.

\begin{table}[b]
\caption{Comparison and descriptions of three formats of label querying.}
\label{tab:conjoint}
\vskip 0.15in
\begin{center}
\begin{small}
\begin{tabular}{llc}
\hline
Format & Information obtained & Mathematical expression \\
\hline
Direct & Continuous valued labels & $\hat{y}^{[i]}, \forall i$ \\
Ranking & Ranking of all labels & $\{i_1, \cdots, i_n \mid \hat{y}^{[i_1]}>\cdots>\hat{y}^{[i_n]}\}$ \\
Choice & Max/min label w/in choice set & $\{i_1^{max},\cdots,i_s^{max} \mid i_j^{max}=\underset{k\in S_k}{\mathrm{argmax}}\   \hat{y}^{[k]}, j\in 1,\cdots,s\},$ \\
&&
$\{i_1^{min},\cdots,i_s^{min} \mid i_j^{min}=\underset{k\in S_k}{\mathrm{argmin}}\   \hat{y}^{[ks]}, j\in 1,\cdots,s\},$ \\
\hline
\end{tabular}
\end{small}
\end{center}
\vskip -0.1in
\end{table}

\subsection{The Choice-Based Questionnaire}
Suppose we have an unlabeled dataset of $n$ examples, $X=\{x^{[1]}, \cdots, x^{[n]}\}, x_i\in\mX$, for which we wish to obtain absolute-scaled, continuous-valued labels suitable for supervised training, $Y=\{y^{[1]}, \cdots, y^{[n]}\}, y_i\in\mY\in\mathbb{R}$. In choice formats, the examples are partitioned in to $n/s$ choice sets (minibatches) each of size $s$. Expert labelers are then asked to choose the most and least risky example within each choice set, $S_k=\{x^{<1>}, \cdots, x^{<s>}\}, x^{<i>}\in X$. See Table \ref{tab:conjoint} for a more rigorous mathematical expression. The superset of all choice sets is called the questionnaire, $Q=\{S_1, \cdots, S_{n/s}\}$. There may be multiple questionnaires $\{Q_1, \cdots, Q_q\}$ assigned to multiple expert labelers; however, each questionnaire consists of same set of examples $X$ and each example appears only once in each questionnaire. The partitioning (or minibatching) of choice sets may be different across different questionnaires to increase the diversity of pairwise comparisons. 

\subsection{From Choice to Score}
In the $l$th questionnaire, we encode the \textit{choice} $c_l^{[i]}$ of example $i$ as follows:

\begin{equation}
    c_l^{[i]}=
    \begin{dcases}
        1, & \text{if } y_l^{<i>}=\max_{j}y_l^{<j>} \\
        -1, & \text{if } y_l^{<i>}=\min_{j}y_l^{<j>} \\
        0, & \text{otherwise}.
    \end{dcases}
\end{equation}

Where $y_l^{<j>}$ are the true labels. We then compute the expected choice by averaging the choices for each example across all $q$ questionnaires

\begin{equation}
\label{eq:encodechoice}
    \bar{c}^{[i]}=\braket{c_l^{[i]}}=\frac{1}{q}\sum_{l=1}^q c_l^{[i]}
\end{equation}

It would be nice to know the functional relationship between choice and the absolute risk score so that we can convert the results of the questionnaire to a standard, absolute-scaled measure of risk. Given a label distribution $P(Y=y)=f(y)$ as a prior, the expected choice can be derived as

\begin{equation}
\label{eq:meanagg}
\begin{split}
    \bar{c}^{[i]} & = \mathbb{E}_{y^{<i>}\sim Y}\enspace c(y^{<i>}) \\
                  & = +1\times P(y^{<i>}=\max_{j}y^{<j>}) \\
                  &\quad -1\times P(y^{<i>}=\min_{j}y^{<j>}) \\
                  &\quad +0\times P(y^{<i>}\text{ is neither max nor min})
\end{split}
\end{equation}

Because of independence (examples are placed into choice sets without regard for the others are), the probability that example $i$ has the largest true label within its choice set is

\begin{equation}
\begin{split}
    P(y^{<i>}=\max_{j}y^{<j>}) &= \prod_{j\neq i}P(y^{<j>}\leq y^{<i>}) \\
      &= \prod_{j\neq i}\left(\int_{-\infty}^{y^{<i>}}f(y^{<j>})\mathrm{d}y^{<j>}\right) \\
      &= \left(\int_{-\infty}^{y^{<i>}}f(y)\mathrm{d}y\right)^{s-1}.
\end{split}
\end{equation}

A similar derivation can be made for the probability that example $i$ has the smallest true label. Thus, a relationship between the choice and risk score which asymptotically approaches convergence in the limit of a large number of questionnaires is

\begin{equation}
    \label{eq:choice2score}
    \bar{c}^{[i]} = \left(\int_{-\infty}^{y^{<i>}}f(y)\mathrm{d}y\right)^{s-1} -                         \left(\int_{y^{<i>}}^{\infty}f(y)\mathrm{d}y\right)^{s-1}.
\end{equation}

There is no analytical inverse for Equation \ref{eq:choice2score}, but the inverse can be readily computed by numerical optimization.

\textbf{Example} Consider a uniform label distribution $f(y)=1/2$ for $y\in [-1,1]$. The expected choice is

\begin{equation}
    \bar{c}^{[i]}=\left(\frac{y}{2}+\frac{1}{2}\right)^{s-1} - \left(-\frac{y}{2}+\frac{1}{2}\right)^{s-1}.
\end{equation}

We conduct an simulation of 25 questionnaires each with the same set of 500 examples whose labels are sampled from a uniform distribution between -1 and 1. In each questionnaire, the examples are randomly partitioned into choice sets of size $s$. An oracle is used to provide the choice $c^{<i>}$ for each example as to whether it is most/least/neither risky its respective choice set. The choices are encoded and mean-aggregated as described by Eqs. \ref{eq:encodechoice}, \ref{eq:meanagg}. Figure \ref{fig:choice2score} plots the averaged choice for each example as function of its true label. The simulation data fits our theoretical result perfectly, verifying that the choice-to-score relation in Eq. \ref{eq:choice2score} can be used to convert \textit{relative} information to \textit{absolute} information about the labels.

\begin{figure}[t] 
\centering
\begin{subfigure}{.24\textwidth}
    \centering
    \includegraphics[width=\textwidth]{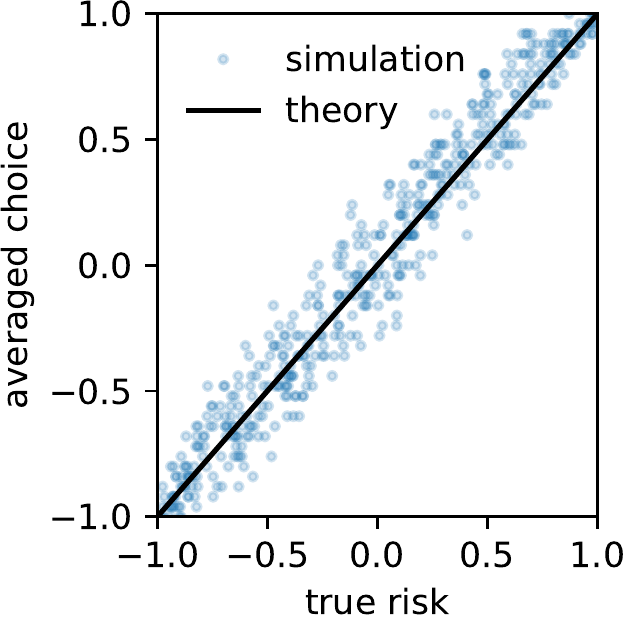}
    \caption{Choice set size = 3}
    \label{fig:setsize3}
\end{subfigure}
\begin{subfigure}{.24\textwidth}
    \centering
    \includegraphics[width=\textwidth]{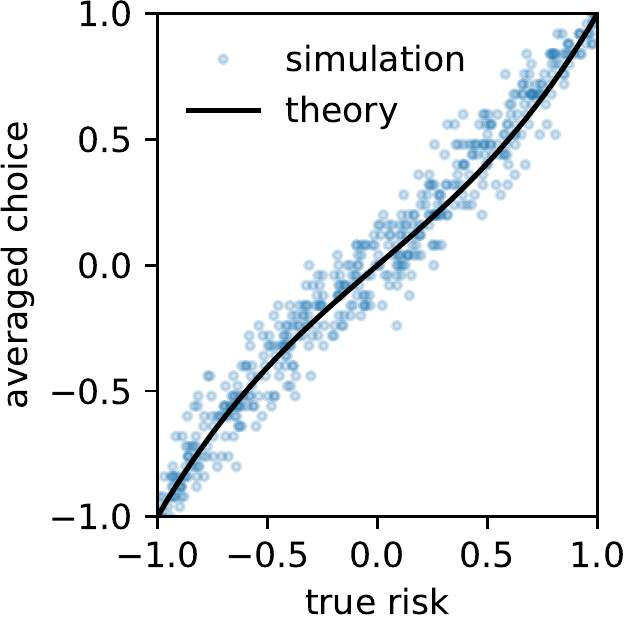}
    \caption{Choice set size = 4}
    \label{fig:setsize4}
\end{subfigure}
\begin{subfigure}{.24\textwidth}
    \centering
    \includegraphics[width=\textwidth]{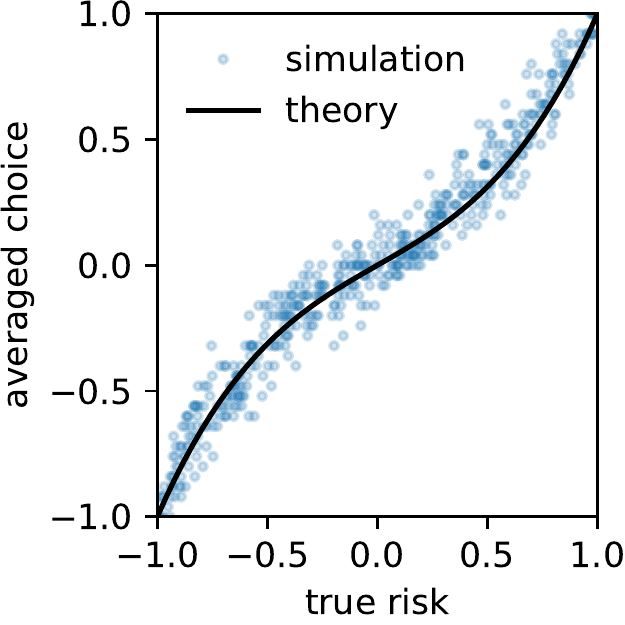}
    \caption{Choice set size = 5}
    \label{fig:setsize5}
\end{subfigure}
\begin{subfigure}{.24\textwidth}
    \centering
    \includegraphics[width=\textwidth]{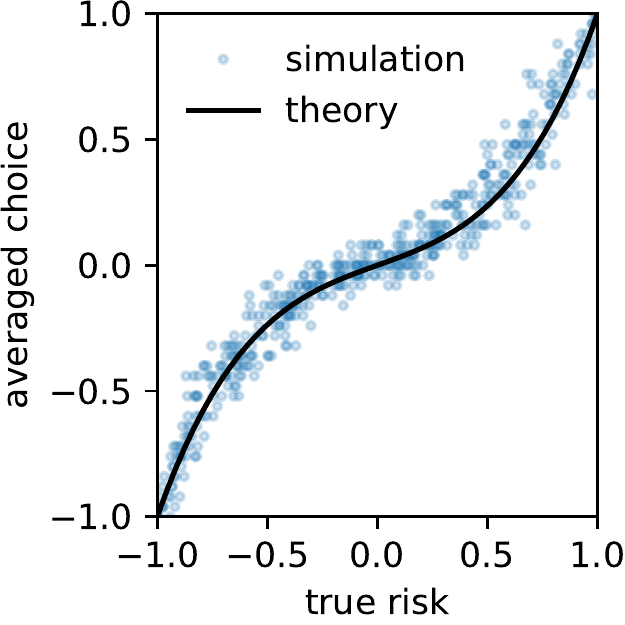}
    \caption{Choice set size = 6}
    \label{fig:setsize6}
\end{subfigure}
\caption{\small Simulation results showing averaged choice as a function of true label for 500 different examples (blue dots). An oracle evaluated each of the examples using the choice-based paradigm over 25 questionnaires. The examples' true labels are sampled from a uniform distribution. \label{fig:choiceexample}}
\end{figure}

\section{End-to-End Algorithm for Inherent Risk Scoring}

\subsection{Overview}
We now present the step-by-step details of the Inherent Risk Scoring Model (IRSM) we built for a large financial institution. The data consisted of investment banking customers during an approved period from 2011 to 2015. All entities considered (financial institutions, sovereign wealth funds, individuals, etc.) were banking participants during that time period. The IRSM was developed with the intended purpose of being used as a risk prioritization framework to quickly estimate the probability that a given customer's profile of only Know Your Customer (KYC) variables was a high or low inherent risk for being involved in money-laudering activity. The IRSM's prediction of a customer's inherent risk based on KYC variables, without access to transaction information, was designed to mimic the kind of information that a fraud investigator, henceforth called Subject Matter Advisor (SMA), has access to when performing a first level money-laundering screening. Each customer (example) consisted of 24 variables (features), which included \ronny{what kind of features were they}. Roughly 156000 unique customers were available.

\subsection{Synthetic Examples via Optimal Experimental Design}

While we had access to many examples (i.e. customer profiles), we did not want our trained model to depend on the particular distribution of our training data, as distributions can vary across different banks. Plus, we did not have the SMA resources to label such a large dataset. To eliminate these issues, we create a synthetic dataset of examples using Optimal Experimental Design (OED). OED operates under the assumption that obtaining a label for each example is costly and presents a theory for selecting examples that, given labels, predict the target variable as well as possible. In particular, we seek to optimize the D-efficiency, defined as $|X^TX|$ where $X\in \mathbb{R}^{n\times m}$ is the matrix of all $n$ examples with $m$ features. Maximizing D-efficiency makes the $n$ examples span the largest possible region in input space, ensuring that the examples are as orthogonal and balanced as possible (\cite{Eriksson2000}).

\begin{figure}
\centering
\includegraphics[width=.45\textwidth]{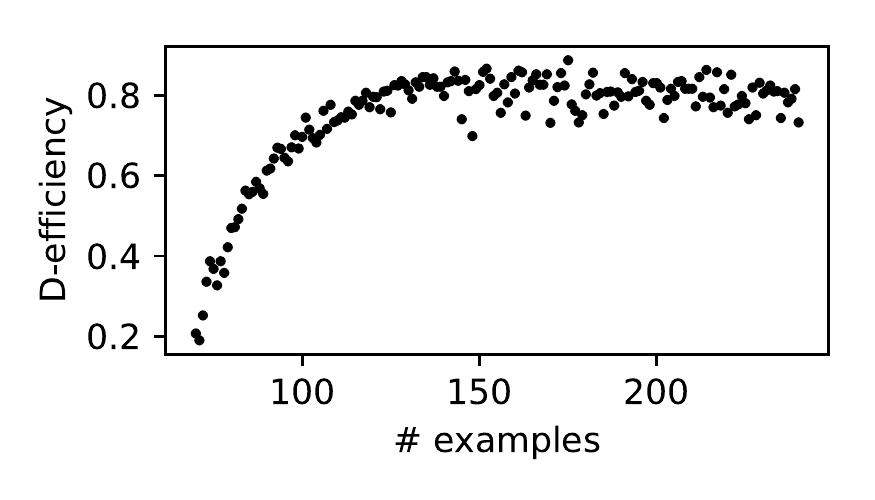}
\caption{\small Optimal D-efficiency computed by the \code{optMonteCarlo} function in R as a function of the number of examples. There are diminishing returns after about 120 examples. \label{fig:deff}}
\end{figure}

We leverage the \code{optMonteCarlo} implementation of the Federov algorithm in the R statistical package for obtaining the optimal design. We observe that the D-efficiency increases with the number of examples allowed up to about 120 examples, after which the gain in D-efficiency diminishes. See Figure \ref{fig:deff}. Based on these results, we chose the number of examples to be $n=188$.

\subsection{Choice Set Size Selection}

\begin{figure}
\centering
\includegraphics[width=.4\textwidth]{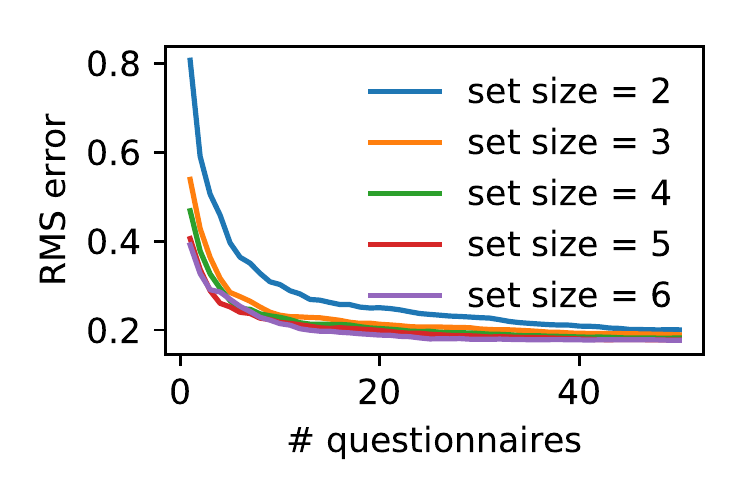}
\caption{\small RMS error between the true label and the label obtained by taking the inverse of Eq. \ref{eq:choice2score} on a set of 188 examples. As the number of questionnaires increases, the error converges to zero. A choice set size of 4 yielded the lowest error at 20 questionnaires, informing our decision of using that size. \label{fig:errorvs}}
\end{figure}

\subsection{Choice Set (Minibatch) Optimization}

\subsection{Crowdsourced Outlier Removal}

\subsection{Model Selection and Training}
Our inherent risk scoring model was trained, validated, and tested on the collected questionnaire results from a team of SMAs (i.e. expert humans). The questionnaires consisted of unique choice sets of synthetic KYC profiles as per the complete methodology outlined in this paper.

The business problem required that the model was tuned to minimize the likelihood of classifying a high risk customer as low risk (i.e. false negatives), compliant with an industry standard false-negative rate of roughly 1:1000.


\begin{figure}[h!t]
\centering
\begin{subfigure}{.4\textwidth}
    \centering
    \includegraphics[width=\textwidth]{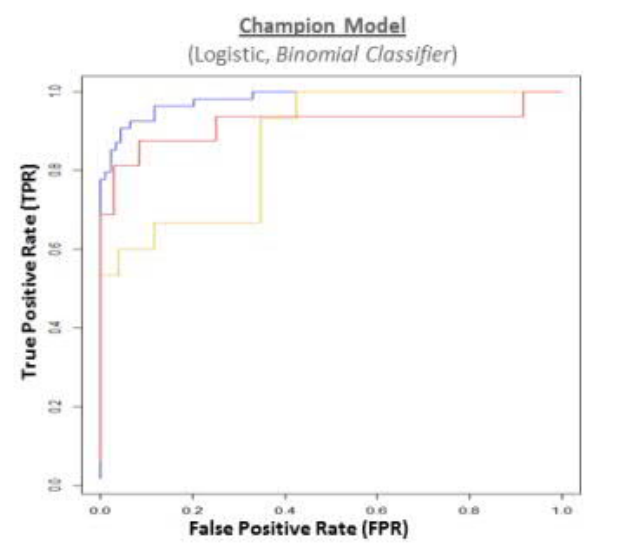}
    \caption{Inherent Risk Model AUC performance on SMA evaluated surveys.}
    \label{fig:auc}
\label{fig:auc_performance}
\begin{tabular}{ |l|l|l|l| }
\hline
\multicolumn{4}{ |c| }{IRM AUC Performance on SMA Evaluated Surveys} \\
\hline
Target & Training AUC & Validation AUC & Test AUC\\ 
\multirow{1}{*}{AUC} & 97\% & 90\% & 93\% \\ 
\multirow{1}{*}{Classification Error} & 8\% & 20\% & 11\% \\
\hline
\end{tabular}

\end{subfigure}
\caption{\small IRM performance on SMA evaluated surveys. Performance on Trainind choice sets is in blue, Validation choice sets in yellow, and Test choice sets in red. See Section \ref{sec:notmadeyet} for more discussion.}
\end{figure}

\subsection{Evaluation}

The review of an account from an SMAs perspective ultimately reduces to whether or not the SMA, in their professional opinion, is legally required to file a Suspicious Activity Report $($SAR$)$ to be filed with the appropriate federal agencies. A simple outline of the process is provided below:

\begin{enumerate}[i]
  \item An alerting system identifies a party's behaviour as breaking some list of statistically defined rules or business logic
  \item The Alert is prioritized and placed in queue for manual review, with the attached criteria that generated the alert
  \item Some prioritization framework decides which alert should be reviewed first, placing it in an SMA queue
  \item The SMA reviews all available information on the alert and the account's activity, making a recommendation for whether or not the alert should be reviewed by a second-level SMA, typically within an area of specialty or with more review experience. If so, it is placed in a second-review queue
  \item The second-review queue provides that alert to a senior SMA. The recommendation of $SAR$ or $Not-SAR$ is then attached to the account and party connected to the alert
\end{enumerate}


The IRM was applied to the customer population and provided a rank-ordering of those profiles. A logic layer stating the presence of one of 14 scenarios associated with risky trading, transaction, or settlement behavior conditioned the sample down to rougly 8000 unique customers. All of these 14 scenarios are captured by the features of the model and were present in the survey construction. The alerting population was further reduced by 4000 if the alert was only generated due to the presence of wash trades, the details of which are out of the scope of this letter. The remaining alerted accounts were individually reviewed providing a rare performance evaluation for the model on real-world data. The SMA-recommendation framework above expected that roughly 4 out of the 4000 alerts may have defeated the SMA-recommendation system prior to the review of the alerting population.

The propritization framework created with the author's complete methodology was tested by letting it select the 1500 profiles it predicted to be the riskiest, partitioning the alerting population.  After EY SMAs reviewed the 4000 profiles and identified which ones should be recommended for escalation, the IRM capability to pick truly "riskier" profiles resulted in a 15.5X improvement in the escalation rate. Compared with the escalation rate associated with a random alerting profile of 0.775\%, the IRM achieved a 2.4X improvement. The authors note that none of the 31 SMA recommended escalations were in the bottom 40\% or the IRM risk score, highlighting a low false negative rate.

\begin{tabular}{ |l|l|l|l| }
\hline
\multicolumn{4}{ |c| }{IRM Performance on Alerted KYC Profiles} \\
\hline
Population Group & KYC Profiles & EY SMA Recommended Escalations & Escalation Rate\\ 
\multirow{1}{*} IRM Selected Alerted Profiles & 1500 & 28 & 1.87\% \\ 
\multirow{1}{*} Remaining Scenario Alerted Profiles & 2500 & 3 & 0.12\% \\
\hline
\end{tabular}

\section{Conclusion}

We have introduced a novel end-to-end methodology for developing machine learning models trained on labels that are defined by relative comparisons in the absence of absolute label definitions. Our IRM model was trained on the collected SMA evaluations from group theoretic optimized surveys that consisted of choice sets of synthetic data profiles using this methodology. The IRM was applied to real-world profiles to evaluate the KYC risk of international banking customers being involved with money-laundering activity achieving a 15.5X improvement of the identification rate of customer profiles that SMAs recommeded for escalation. These results challenge the data science community to extend the class of models that can be built and deployed to solve applied mathematics and economics problems, by first re-evaluating the ground truth representation of the information and states being modeled, and then extending the semantic interpretation of those models by connecting them to a human-machine interface before beginning the model building process. We propose a general class of solution that applies information theoretic tools to a novel form of crowdsourcing which enables the extension of expert systems to classificaiton tasks without absolute label definitions.


\section{Related work}
\cite{Xu2018}
\cite{Ratner2016}

\subsubsection*{Acknowledgments}

\bibliography{nips_2017}